\documentclass[letterpaper]{article}
\usepackage{aaai}
\usepackage{times}
\usepackage{helvet}
\usepackage{courier}

\usepackage{graphicx}
\usepackage{amsmath,amssymb}
\usepackage{subfigure}
\usepackage{array}
\usepackage{multirow}
\usepackage{url}
\usepackage{natbib}

\newcolumntype{V}{>{$\vcenter\bgroup\hbox\bgroup}c<{\egroup\egroup$}}

\graphicspath{{./imgs/}}

\begin{document}
%
\title{Sample-Targeted Clinical Trial Adaptation\\[5pt]}
\author{Ognjen Arandjelovi\'c\\[5pt]Centre for Pattern Recognition and Data Analytics, Deakin University, Australia}
\maketitle
\begin{abstract}
\begin{quote}
Clinical trial adaptation refers to any adjustment of the trial protocol after the onset of the trial. The main goal is to make the process of introducing new medical interventions to patients more efficient by reducing the cost and the time associated with evaluating their safety and efficacy. The principal question is how should adaptation be performed so as to minimize the chance of distorting the outcome of the trial. We propose a novel method for achieving this. Unlike previous work our approach focuses on trial adaptation by sample size adjustment. We adopt a recently proposed stratification framework based on collected auxiliary data and show that this information together with the primary measured variables can be used to make a probabilistically informed choice of the particular sub-group a sample should be removed from. Experiments on simulated data are used to illustrate the effectiveness of our method and its application in practice.
\end{quote}
\end{abstract}

\section{Introduction}
Robust evaluation is a crucial component in the process of introducing new medical interventions. Amongst others, these include newly developed medications, novel means of administering known treatments, new screening procedures, diagnostic methodologies, physio-therapeutical manipulations, and many others. Such evaluations usually take on the form of a controlled clinical trial (or a series thereof), the framework widely accepted as best suited for a rigourous statistical analysis of the effects of interest~\citep{Mein1986,Pian1997,FrieFurbDeMe1998} (for a related discussion and critique also see \citep{Pens2005}). Driven both by legislating bodies, as well as the scientific community and the public, the standards that the assessment of novel interventions are expected to meet continue to rise. Generally, this necessitates trials which employ larger sample sizes and which perform assessment over longer periods of time. A series of practical challenges emerge as a consequence. Increasing the number of individuals in a trial can be difficult because some trials necessitate that participants meet specific criteria; volunteers are also less likely to commit to participation over extended periods of time. The financial impact is another major issue -- both the increase in the duration of a trial and the number of participants result in additional cost to an already expensive process. In response to these challenges, the use of adaptive trials has emerged as a potential solution~\citep{Fish1998,_US2010,HungWangONei2006}. The key idea underlying the concept of an adaptive trial design is that instead of fixing the parameters of a trial before its onset, greater efficiency can be achieved by adjusting them as the trial progresses~\citep{ChowChan2011}. For example, the trial sample size (e.g.\ the number of participants in a trial), treatment dose or frequency, or the duration of the trial may be increased or decreased depending on the accumulated evidence~\citep{CuiHungWang1999,Niss2006,Lang2011}.

\vspace{-10pt}\paragraph{Method overview}
The method for trial adaptation we describe in this paper has been influenced by recent work on the analysis of imperfectly blinded clinical trials~\citep{Aran2012g,Aran2012h}. Its key contribution was to introduce the idea of trial outcome analysis by patient sub-groups which comprise trial participants matched by the administered intervention (treatment or control) and their responses to an auxiliary questionnaire in which the participants are asked to express their belief regarding their assignment intervention in the closed-form. This framework was shown to be suitable for robust inference in the presence of ``unblinding'' \citep{Aran2012g,HaahHrob2006}. The method proposed in the present paper emerges from the realization that the same framework can be used for trial adaptation by providing information which can be used to make a statistically informed selection of the trial participants which can be dropped from the trial before its completion, without significantly affecting the trial outcome. Thus, the proposed approach falls under the category of trial adaptations by ``amending sample size'', in contrast to ``dose finding'' or ``response adapting'' methods which dominate previous work~\citep{Lang2011}.

In~\citep{Aran2012g} it was shown that the analysis of a trial's outcome should be performed by aggregating evidence provided by matched participant sub-groups, where two sub-groups are matched if they contain participants who were administered different interventions but nonetheless had the same responses in the auxiliary questionnaire. Therefore, our idea advanced here is that an informed trial sample size reduction can be made by computing which matched sub-group pair's contribution of useful information is affected the least with the removal of participants from one of its groups.

\vspace{-10pt}\paragraph{Contrast with previous work}
Before introducing the proposed method in detail, it is worthwhile emphasizing two fundamental aspects in which it differs from the methods previously described in the literature. The first difference concerns the nature of the statistical framework which underlies our approach. Most of the existing work on trial adaptation by sample size adjustment adopts the frequentist paradigm. These methods follow a common pattern: a particular null hypothesis is formulated which is then rejected or accepted using a suitable statistic and the desired confidence requirement~\citep{JennTurn2003}. In contrast, the method described in this paper is thoroughly Bayesian in nature. The second major conceptual novelty of the proposed method lies in the question it seeks to answer. All previous work on trial adaptation by sample size adjustment addresses the question of \emph{whether} the sample size can be reduced while maintaining a certain level of statistical significance of the trial's outcome. In contrast, the present work is the first to ask a complementary question of \emph{which} particular individuals in the sample should be removed from the trial once the decision of sample size reduction has been made. Thus, the proposed method should not be seen as an alternative to the any of the previously proposed methods but rather as a complementary element of the same framework.

\subsection{Auxiliary data collection}
\label{ss:auxData}
The type of auxiliary data collection we utilize in this work was originally proposed for the assessment of blinding in clinical trials~\citep{JameBlocLeeKraeFull1996}. Since then it has been adopted for the same purpose in a number of subsequent works~\citep{BangNiDavi2004,HrobForfHaahAlsN+2007,KolaBangPark2009,Aran2012g} (also see~\citep{Sack2007} for related commentary). The questionnaire allows the trial participants to express their belief on the nature of the intervention they have been administered (control or treatment) using a fixed number of choices. The most commonly used, coarse-grained questionnaire admits the following three choices:
\begin{list}{test}{\leftmargin=15pt}
  \item[1:] belief that control intervention was administered,\vspace{-3pt}
  \item[2:] belief that treatment intervention was administered, and\vspace{-3pt}
  \item[3:] undecidedness about the nature of the intervention.
\end{list}

\subsection{Matching sub-groups outcome model}
\label{ss:subgroups}
In the general case, the effectiveness of a particular intervention in a trial participant depends on the inherent effects of the intervention, as well as the participant's expectations (conscious or not). Thus, as in~\citep{Aran2012g}, in the interpretation of trial results, we separately consider each population of participants which share the same combination of the type of intervention and the expressed belief regarding this group assignment. For example, when a 3-tier questionnaire is used in a trial comparing the administration of the treatment of interest and control, we recognize control sub-groups:
\begin{list}{test}{\leftmargin=15pt}
\item[$G_{C-}$:] participants of the control group who believe they were assigned to the control group,\vspace{-3pt}
\item[$G_{C0}$:] participants of the control group who are unsure of their group assignment,\vspace{-3pt}
\item[$G_{C+}$:] participants of the control group who believe they were assigned to the treatment group,
\end{list}
and the three corresponding treatment sub-groups. The key idea underlying the method proposed in~\citep{Aran2012g} is that because the outcome of an intervention depends on both the inherent effects of the intervention and the participants' expectations, the effectiveness should be inferred in a like-for-like fashion. In other words, the response observed in, say, the sub-group of participants assigned to the control group whose feedback professes belief in the control group assignment should be compared with the response of only the sub-group of the treatment group who equally professed belief in the control group assignment.


\section{Sub-group selection}
\label{s:prop}
The primary aim of the statistical framework described in~\citep{Aran2012g} is to facilitate an analysis of trial data robust to the presence of partial or full unblinding of patients, or indeed patient preconceptions which too may affect the measured outcomes. Herein we propose to exploit and extend this framework to guide the choice of which patients are removed from the trial after its onset, in a manner which minimizes the loss of statistical significance of the ultimate outcomes.

At the onset of the trial, the trial should be randomized according to the current best clinical practice; this problem is comprehensively covered in the influential work by \cite{Berg2005}. If a reduction in the number of trial participants was attempted at this stage, by the very definition of a properly randomized trial, statistically speaking there is no reason to prefer the removal of any particular subject (or indeed a set of subjects) over another. Instead, any trial size adaptation must be performed at a later stage after some meaningful differentiation between subjects takes place~\citep{Nels2010}.

The most obvious observable differentiation that takes place between patients as the trial progresses is that of the outcomes of primary interest in the trial (the ``response''). This differentiation may allow for a statistically informed choice to be made about which trial participants can be dropped from the trial in a manner which minimizes the expected distortion of the ultimate findings. For example, this can be done by seeking to preserve the distribution of measured outcomes within a group (treatment or control) but with the constraint of a smaller number of participants; indeed, our approach partially exploits this idea. However, our key contribution lies in a more innovative approach, which exploits additional, yet readily collected discriminative information. The proposed approach not only minimizes the effect of smaller participant groups but also ensures that no unintentional bias is injected due to imperfect blinding. Recall that the problem of inference robust to imperfect blinding should always be considered, as blinding can only be attempted with respect to those variables of the trial which have been identified as revealing of the administered treatment (and even for these it is fundamentally impossible to \emph{ensure} perfect blinding).

\begin{figure}
  \centering
  \includegraphics[width=0.3\textwidth]{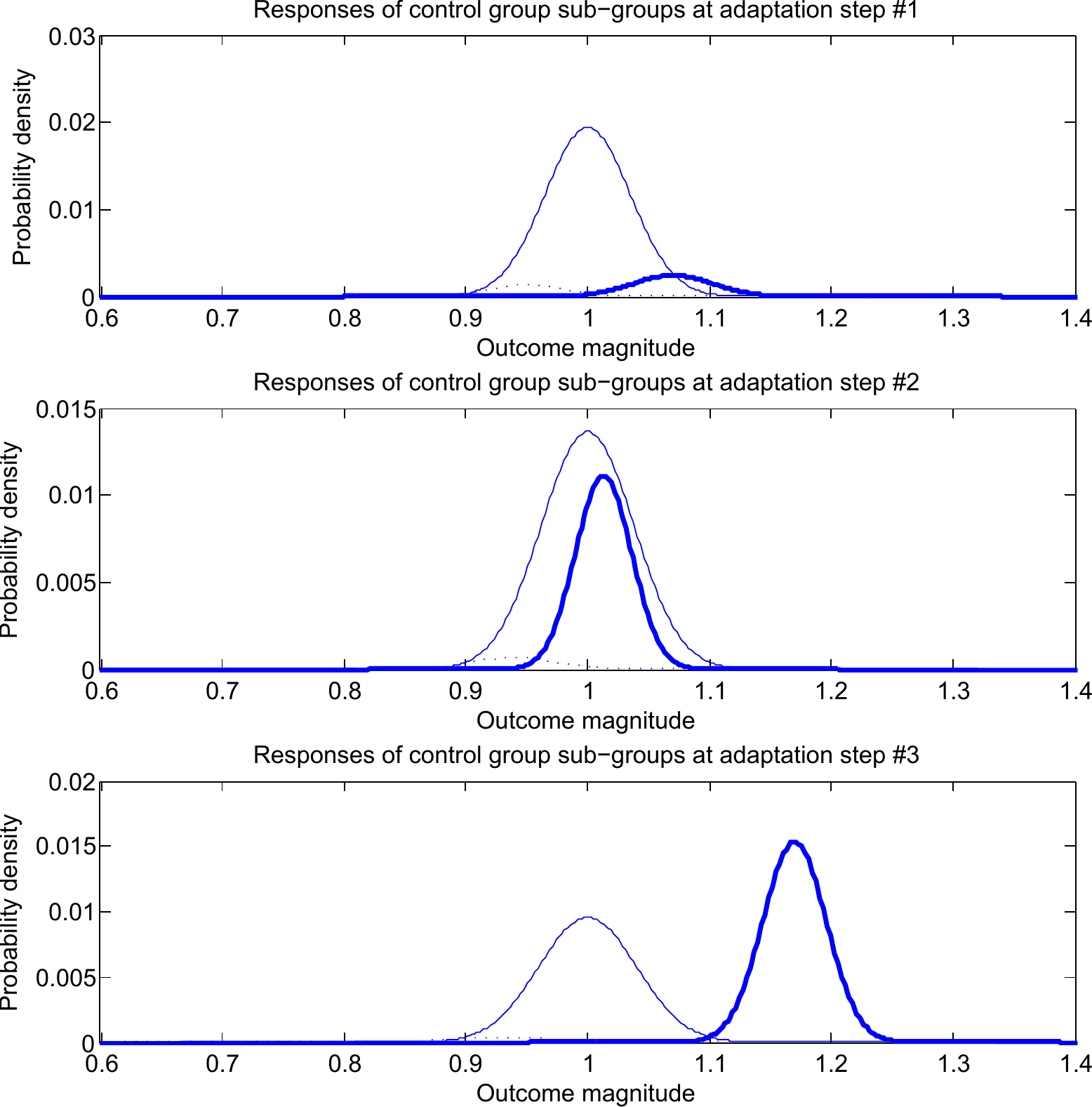}
  \caption{\small A conceptual illustration on a hypothetical example of the phenomenon whereby trial participants change their sub-group membership (recall that each sub-group is defined by its members' intervention assignment \emph{and} auxiliary questionnaire responses). This is quite likely to occur when the effects of the treatment are very readily apparent but various other mechanisms can act so as to cause a non-zero and changing sub-group flux.  }
  \label{f:progress}
  \vspace{-20pt}
\end{figure}

Our idea is to administer an auxiliary questionnaire of the form described in~\citep{JameBlocLeeKraeFull1996,BangNiDavi2004} every time an adaptation of the trial group size is sought. As in~\citep{Aran2012g}, this leads to the differentiation of each group of participants (control or treatment) into sub-groups, based on their belief regarding their group assignment. In general, this means that even if no participants are removed from the trial, a participant may change his/her sub-group membership status. This is illustrated with a hypothetical example in Fig.~\ref{f:progress}. The first time an auxiliary questionnaire is administered (top plot), most of the treatment group participants are still unsure of their assignment (solid blue line); a smaller number of participants have correctly guessed (or inferred) their assignment (bold blue line); lastly, an even smaller number holds the incorrect belief that they are in fact members of the control group (dotted blue line). All of the sub-groups show a spread of responses to the treatment, such as may be expected due to various personal variations of their members. At the time of the second snapshot (middle plot), at the next instance when auxiliary data is collected, the proportions of participants in each sub-group has changed, as do the associated treatment response statistics. A similar observation can be made with respect to the third and the last snapshot pictured in the figure (bottom plot). This sort of a development would not be unexpected -- if the treatment is effective, as the trial progresses there will be an increase in the number of treatment group participants who observe and correctly interpret these changes (note that this also means that there will be an associated increase in the number of participants who may exhibit an additional positive effect from the fortunate realization that they are receiving the studied treatment intervention, rather than the control intervention). That being said, it should be emphasized that no assumption on the statistics of sub-group memberships or their relative sizes is made in the proposed method. The example in Fig.~\ref{f:progress} is merely used for illustration.

The question is: how does this differentiation of patients by auxiliary data sub-groups help us make a statistically robust choice of which participants in the trial should be preferentially dropped if a reduction in the trial size is sought? To answer this question, recall that the main premise of~\citep{Aran2012g} is``that it is meaningful to compare only the corresponding treatment and control participant sub-groups, that is, sub-groups matched by their auxiliary responses.'' Each sub-group comparison contributes information used to infer the probability density of the differential effects of the treatment. We can then reformulate the original question as: from which matching sub-group pair should participants be preferentially dismissed from further consideration so as to best conserve the sub-group pair's information contribution? Consider how the information on the differential effects between a single pair of matching sub-groups is inferred. In its general form, we can estimate some distance between the distributions of the two sub-groups using a Bayesian approach:
{\small\begin{align}
  \rho^* \propto &\int_{\Theta_c} \int_{\Theta_t}
  \underbrace{\rho(p_c(x;\Theta_c),p_t(x;\Theta_t))}_{\tiny \begin{array}{c}
                                                  \text{Distance between distributions}\\
                                                  \text{for specific parameter values}\\
                                                \end{array}}\times \notag\\
  &\underbrace{p(D_c|\Theta_c)~p(D_t|\Theta_t)}_{\text{Model likelihoods}}
  \underbrace{p(\Theta_c)~p(\Theta_t)}_{\text{Parameter priors}} \times ~d\Theta_t~d\Theta_c
  \label{e:bayes}
\end{align}}
where {\small$\Theta_c$} and {\small$\Theta_t$} are the sets of variables parameterizing the two corresponding distributions {\small$p_c(x;\Theta_c)$} and {\small$p_t(x;\Theta_t)$}, {\small$p(\Theta_c)$} and {\small$p(\Theta_t)$} the parameter priors, {\small$\rho(p_c(x;\Theta_c),p_t(x;\Theta_t))$} a particular distance function (e.g.\ the Kullback-Leibler divergence~\citep{KullLeib1951}, Bhattacharyya~\citep{Bhat1943} or Hellinger distances~\citep{Hell1909}, or indeed the posterior of the difference of means used in~\citep{Aran2012g}), and {\small$D_c$} and {\small$D_t$} the measured trial outcomes (e.g.\ the reduction in blood plasma LDL in a statin trial etc).

Note that by changing (reducing) the number of participants in one of the groups, the only affected term on the right hand side of \eqref{e:bayes} is one of the likelihood terms, {\small$p(D_c|\Theta_c)$} or {\small$p(D_t|\Theta_t)$}. Seen another way, a change in the number of participants in the trial changes the \emph{weighting} of the product of the distance term {\small$\rho(p_c(x;\Theta_c),p_t(x;\Theta_t))$} and the priors {\small$p(\Theta_c)~p(\Theta_t)$}. Our idea is then to choose to remove a trial participant from that sub-group which produces the smallest change in the estimate {\small$\rho^*$}. However, it is not clear how this may be achieved, since it is the size of the set {\small$D_c$} that is changing (so, for example, treating {\small$D_c$} and {\small$D_t$} as vectors and {\small$f$} as a function of vectors would not achieve the desired aim). Examining the sensitivity of {\small$\rho*$} with the removal of each datum (i.e.\ trial participant) from {\small$D_c$} and {\small$D_t$} is also unsatisfactory since the problem does not lend itself to a greedy strategy: the optimal choice of which {\small$n_{rem}$} trial participants to drop from the trial cannot be made by making {\small$n_{rem}$} optimal choices of which \emph{one} participant to drop. An approach following this direction but attempting to examine all possible sets of size {\small$n_{rem}$} would encounter computational tractability obstacles since this problem is NP-complete. The alternative which we propose is to consider and compare the magnitudes of partial derivatives of {\small$\rho^*$} with respect to the sizes of data sets {\small$D_c$} and {\small$D_t$}, but with an important constraint -- the derivatives are taken of the \emph{expected} functional form of {\small$\rho^*$} over different members of {\small$D_c$} and {\small$D_t$}. Formalizing this, we compute:
{\small\begin{align}
  E\left[\frac{\partial \rho^*}{\partial n_c}\right]_{D_c} &&\text{ and }&&
  E\left[\frac{\partial \rho^*}{\partial n_t}\right]_{D_t},
  \label{e:diff}
\end{align}}
where {\small$E[\rho^*]_{D_c}$} and {\small$E[\rho^*]_{D_t}$} are respectively the expected values of {\small$\rho^*$} across the space of possible observations in {\small$D_c$} and {\small$D_t$}. Thus {\small$E[\rho^*]_{D_c}$} and {\small$E[\rho^*]_{D_t}$} are functions of two scalars, the sizes {\small$n_c$} and {\small$n_t$} of sets {\small$D_c$} and {\small$D_t$} i.e.\ the numbers of members of the corresponding sub-groups.

The proposed solution is not only theoretically justified but it also lends itself to simple and efficient implementation. Since the expected values {\small$E[\rho^*]_{D_c}$} and {\small$E[\rho^*]_{D_t}$} are evaluated over sets {\small$D_c$} and {\small$D_t$}, in \eqref{e:bayes} the only term affected is {\small$p(D_c|\Theta_c)~p(D_t|\Theta_t)$}, so the solution is readily obtained as a closed form expression. Equally, the integration is readily performed using one of the standard Markov chain Monte Carlo integration methods~\citep{Gilk1995}.

\section{Application example}
\label{ss:example}
In order to illustrate how the described method could be applied in practice, let us consider a hypothetical example. Let the trial observation data in two matching sub-groups be drawn from the random variables {\small$X_c$} and {\small$X_t$}, which are appropriately modelled using normal distributions~\citep{AitcBrow1957}:
{\small$X_t \sim 1/\sigma_t \exp -(x - m_t)^2/(2\sigma_t^2)$} and {\small$X_c \sim 1/\sigma_c \exp -(x - m_c)^2/(2\sigma_c^2)$}. The next step is to choose an appropriate distance function {\small$\rho$} in \eqref{e:bayes}. In practice, this choice would be governed by the goals of the study. Herein, for illustrative purposes we choose {\small$\rho$} to be the probability that a patient will do better when the treatment intervention is administered:
{\small\begin{align}
    \rho(p_t(x;\Theta_t),&p_c(y;\Theta_c))=
    \int_0^\infty \int_0^x p_t(x;\Theta_t)~p_c(y;\Theta_c)~dy~dx \notag
\end{align}}
where {\small$\Theta_c=(m_c,\sigma_c)$} and {\small$\Theta_t=(m_t,\sigma_t)$} are the mean and standard deviation parameters specifying the corresponding normal distributions.
{\small\begin{align}
    \rho^*\propto
     \int_0^\infty \int_{-\infty}^\infty \int_0^\infty &\int_{-\infty}^\infty \int_0^\infty \int_0^x&  \notag\\
                        p_t(x | m_t, \sigma_t)~&p(m_t, \sigma_t)~dx~dm_t~d\sigma_t ~\times \notag\\
                        p_c(y | m_c, \sigma_c)~&p(m_c, \sigma_c)~dy~dm_c~d\sigma_c \notag\\
                        =&\int_0^\infty \int_0^x I_t(x)~I_c(y)~dy~dx
     \label{e:integral}
\end{align}}
where -- assuming uninformed priors on {\small$m_c$}, {\small$m_t$}, {\small$\sigma_c$}, and {\small$\sigma_t$} -- each of the integrals {\small$I_t(x)$} and {\small$I_c(y)$} has the form:
{\small\begin{align}
  I = \int_0^\infty \int_{-\infty}^\infty &\frac{1}{\sigma}~\exp\left\{ -\frac{(x-m)^2}{2\sigma^2} \right\}~ \notag\\
  \times~&\frac{1}{\sigma^n}~\exp\left\{ -\frac{\sum_{i=1}^n (x_i-m)^2}{2\sigma^2} \right\}~dm~d\sigma,
  \label{e:I}
\end{align}}
and {\small$\{x_i\}$} and {\small$n$} stand for either {\small$\{x^{(c)}_i\}$} and {\small$n_c$} or {\small$\{x^{(t)}_i\}$} and {\small$n_t$}, and {\small$\{x^{(c)}_i\}$} and {\small$\{x^{(t)}_i\}$} ($i = 1\ldots n_t$) are exponentially transformed measured trial variables. This integral can be evaluated by combining the two exponential terms and completing the square of the numerator of the exponent as in~\citep{Aran2012g} which leads to the following simplification of \eqref{e:integral}:
{\small\begin{align}
  I &\propto \int_0^\infty \frac{1}{\sigma^{n+1}}~\exp\left\{-\frac{c}{2\sigma^2}\right\}~d\sigma,
  \label{e:intGamma}
\end{align}}
where the value of the only non-constant term is {\small$c = x^2 + \sum_{i=1}^n {x_i}^2 - (x+\sum_{i=1}^n x_i)^2/(n+1)$}.
The form of the integrand in \eqref{e:intGamma} matches that of the inverse gamma distribution {\small$\text{Gamma}(z; \alpha,\beta)=\frac{\beta^\alpha}{\Gamma(\alpha)}~z^{-\alpha-1}~\exp\{-\beta/z\}$}.
The variable {\small$z$} and the two parameters of the distribution, {\small$\alpha$} and {\small$\beta$}, can be matched with the terms in \eqref{e:intGamma} and the density integrated out, leaving:
{\small\begin{align}
  I \propto c^{-\frac{n-1}{2}}.
  \label{e:intermed1}
\end{align}}
Remembering that the functional form of {\small$c$} is different for the control and the trial groups (since it is dependent on {\small$x_i$} which stands for either {\small$x^{(c)}_i$} or {\small$x^{(t)}_i$}), and substituting the result from \eqref{e:intermed1} back into \eqref{e:integral} gives the following expression for the distance function:
{\small\begin{align}
  \rho^*&=\int_0^\infty \int_0^x p_t(x)~p_c(y)~dy~dx
  \propto \int_0^\infty c_t^{-\frac{n_t-1}{2}}\int_0^x  c_c^{-\frac{n_c-1}{2}}~dy~dx\notag
\end{align}}
Our goal now is to evaluate {\small$S_c(\rho^*)$} and {\small$S_t(\rho^*)$}, the sensitivities of the distance function to the change in the size of the control and the treatment groups. Without loss of generality, let us consider {\small$S_t(\rho^*)$}:
{\small\begin{align}
  S_t(\rho^*&) \propto \int_{-\infty}^\infty \int_{-\infty}^x S[I_t(x)]~I_c(y)~dy~dx
  \label{e:St}
\end{align}}
To evaluate {\small$S[I_t(x)]$} we will employ the standard chain rule and perform differentiation with respect to {\small$n_t$} when the corresponding term is a function of the number of treatment participants but not any {\small$x^{(t)}_i$}. On the other hand, as described previously, to handle those terms which do depend on {\small$x^{(t)}_i$} (through {\small$c_t$}), we will use the expected value of the change in the term, averaged over all possible {\small$x^{(t)}_i$} that a unitary decrease in {\small$n_t$} can be achieved. Applying this idea on the expression in \eqref{e:I}:
{\small\begin{align}
  S[&I_t(x)] = -\int_0^\infty \int_{-\infty}^\infty \frac{1}{\sigma}\exp\left\{-\frac{(x-m)^2}{2\sigma^2}\right\} \times \notag\\
   &\Bigg[\frac{\ln\sigma}{\sigma^n}\exp\left\{-\frac{\sum_{i=1}^n (x_i-m)^2}{2\sigma^2}\right\} +\notag\\
   &\frac{1}{\sigma^n}~\exp\left\{-\frac{\sum_{i=1}^n (x_i-m)^2}{2\sigma^2}\right\}~\frac{\sum_{i=1}^n (x_i-m)^2}{2\sigma^2 n^2} \Bigg]~dm~d\sigma
   \label{e:SIt}
\end{align}}
noting that we used the standard result {\small$\frac{d}{dn} \frac{1}{\sigma^n}=-\frac{\ln \sigma}{\sigma^n}$} without including its derivation with intermediary steps shown explicitly. Full double integration in \eqref{e:SIt} is difficult to perform analytically. However, one level of integration -- with respect to {\small$m$} -- is readily achieved. Note that the first term, as a function of {\small$m$}, has the same form as the integral in \eqref{e:I} which we already evaluated. The same procedure which uses the completion of the square in the exponential term can be applied here as well (note that unlike in \eqref{e:I} here it is important to keep track of the multiplicative constants as these will be different for the second term in \eqref{e:SIt}). The integrand in the second term can be expressed in the form {\small$\propto (z-\lambda)^2~\exp -z^2~dz$}. This integration is also readily performed using the standard results
{\small$\int_{-\infty}^{\infty} \frac{1}{\sqrt{2\pi}}~z^2~\exp -\frac{z^2}{2}~dz = 1$} and {\small$\int_{-\infty}^{\infty} \frac{1}{\sqrt{2\pi}}~\exp -\frac{z^2}{2}~dz = 1$}
and by noting that the integrand is an odd function: {\small$\int_{-\infty}^{\infty} \frac{1}{\sqrt{2\pi}}~z~\exp -\frac{z^2}{2}~dz = 0$}. A straightforward application to \eqref{e:SIt} leads to the following expression for the sensitivity {\small$S[I_t(x)]$} of the integral {\small$I_t$} to changes in the size of the corresponding sub-group:
{\small\begin{align}
  S[I_t&(x)] = -\int_0^\infty \frac{\ln \sigma}{\sigma^{n+1}}~\frac{\sigma}{a}~\sqrt{2\pi}~
                \exp\left\{-\frac{c}{2\sigma^2}\right\}~d\sigma \\-
               &\int_0^\infty \frac{\exp\left\{-\frac{c}{2\sigma^2}\right\}}{2\sigma^{n+3} n^2}~
  \left[ n\left(\frac{\sigma}{a}\right)^3 +\sqrt{2\pi}\frac{\sigma}{a} \sum_{i=1}^n x_i \right]~d\sigma \notag
\end{align}}
This result, together with the expression in \eqref{e:intermed1}, can be substituted into \eqref{e:St} and the remaining integration performed numerically.

\subsection{From target sub-groups to specific participants}
Adopting the framework proposed in~\citep{Aran2012g} whereby the analysis of a trial takes into account sub-groups of trial participants, which emerge from grouping participants according to their assigned intervention and auxiliary data, thus far we focused on the problem of choosing the sub-group from which participants should be preferentially removed if a reduction in trial size is sought. The other question which needs to be considered is how \emph{specific} sub-group members are to be chosen, once the target sub-group is identified. Fortunately, the proposed framework makes this a simple task. Recall that the observed trial data within each sub-group is assumed to comprise an identically and independently distributed sample from the underlying distribution, i.e.\ {\small$x_i^{(c)} \sim X_c$} and {\small$x_i^{(t)} \sim X_t$}. This means that it is sufficient to randomly sample the set of target sub-group members to select those which can be removed.

The simplicity of the selection process that our approach allows has an additional welcome consequence. Recall that in the proposed method the choice of the target sub-group is made by comparing differentials in \eqref{e:diff}. It is important to observe that their values are computed for the initial values of {\small$n_c$} and {\small$n_t$}. Thus, as the number of participants in either of the sub-group is changed, so do the values of the differentials, and thus possibly the optimal sub-group choice. This is why the removal of participants should proceed sequentially.

\section{Evaluation}
The primary novelty introduced in this paper is of a methodological nature. In the previous section we explained in detail the mathematical process involved in applying the proposed methodology in practice. Pertinent results were derived for a specific distance function used to quantify the difference in the outcomes between the control and treatment groups in a trial. The choice of the distance function -- which would in practice be made by the clinicians to suit the aims of a specific trial -- governs the relative loss of information when participants are removed from a specific sub-group, and consequently dictates the choice of the optimal sub-group from which the removal should be performed if the overall trial sample size needs to be reduced.

In this section we apply the derived results on experimental data, and evaluate and discuss the performance of the proposed methodology. We adopt the evaluation protocol standard in the domain of adaptive trials research, and obtain data using a simulated experiment.

\subsection{Experimental setup}
\label{ss:setup}
We simulated a trial involving 180 individuals, half of which were assigned to the control and the other half to the treatment group. For each individual we maintain a variable which describes that person's belief regarding his/her group assignment. Thus, for the control group we have {\small$n_c$} beliefs {\small$b^{(c)}_i$ ($i=1\ldots n_c$)} and similarly for the treatment group {\small$n_t$} beliefs {\small$b^{(t)}_i$ ($i=1\ldots n_t$)}. Belief is expressed by a real number, {\small$\forall i.~b^{(c)}_i, b^{(t)}_i \in (-\infty,+\infty)$}, with 0 indicating true undecidedness. Negative beliefs express a preference towards the belief in control group assignment, and positive towards the belief in treatment group assignment. The greater the absolute value of a belief variable is, the greater is the person's conviction. We employ a three-tier questionnaire. To simulate a participant's response, we map the corresponding belief to one of the three possible questionnaire responses according to the following thresholding rule:
{\small\begin{align}
  b < -1           &\rightarrow \text{Belief in control group assignment}\\
  -1 \leq b \leq 1 &\rightarrow \text{Uncertain (``don't know'')}\\
  1 < b            &\rightarrow \text{Belief in treatment group assignment}
\end{align}}
The starting beliefs of participants, i.e.\ their beliefs before the onset of the trial, are initialized to:
{\small\begin{align}
  b^{(c)}_i = b^{(t)}_i=
    \begin{cases}
      -1 \text{ for } i=1 \ldots 9\\
      \hspace{8pt}0 \text{ for } i=10 \ldots 81\\
      \hspace{8pt}1 \text{ for } i=82 \ldots 90
    \end{cases}
\end{align}}
This initialization models the conservative belief of most individuals, and the tendency of a smaller number of individuals to exhibit ``pessimistic'' or ``optimistic'' expectations. The same distribution was used both for the control and the treatment groups, reflecting a well performed randomization.

\vspace{-10pt}\paragraph{Effect accumulation}
As the trial progresses the effects of the treatment accumulate. These are modelled as positive i.e.\ the treatment is modelled as successful in the sense that on average it produces a superior outcome in comparison with the control intervention. We model this using a stochastic process which captures the variability in participants' responses to the same treatment. Specifically, at the discrete time step {\small$k+1$} (the onset of the trial corresponding to {\small$k=0$}), the effects on the {\small$i$}-th treatment and control group participants at the preceding time step {\small$k$} are updated as:
{\small\begin{align}
  e^{(t)}_i(k+1)=e^{(t)}_i(k) + w^{(t)}_i(k+1) \times \exp\left\{-\frac{k+1}{10}\right\}\\
  e^{(c)}_i(k+1)=e^{(c)}_i(k) + w^{(c)}_i(k+1) \times \exp\left\{-\frac{k+1}{10}\right\}
\end{align}}

where {\small$w^{(t)}_i(k+1)$} and {\small$w^{(c)}_i(k+1)$} are drawn from {\small$W_t \sim \mathcal{N}(0.02,0.05)$} and {\small$W_c \sim \mathcal{N}(0.00,0.05)$} respectively. At the onset there is no effect of the treatment; thus:
{\small\begin{align}
  \forall.i=1\ldots n_t.~e^{(t)}_i(0)=0 &&\text{ and }&&
  \forall.i=1\ldots n_c.~e^{(c)}_i(0)=0 \notag
\end{align}}

\vspace{-20pt}\paragraph{Belief refinement}
As the effects of the respective interventions are exhibited, the trial participants have increasing amounts of evidence available guiding them towards forming the correct belief regarding their group assignment. In our experiment this process is also modelled using a stochastic process which is dependent on the magnitude of the effect that an intervention has in a particular participant, as well as uncertainty and differences in people's inference from observations. At the time step {\small$k+1$}, the beliefs of the {\small$i$}-th treatment and control group participants at the preceding time step {\small$k$} are updated as follows:
{\small\begin{align}  
  b^{(t)}_i(k+1)=b^{(t)}_i(k) + 0.01 ~ e^{(t)}_i(k+1) + \omega^{(t)}_i(k+1)\\
  b^{(c)}_i(k+1)=b^{(c)}_i(k) + 0.01 ~ e^{(c)}_i(k+1) + \omega^{(c)}_i(k+1)
\end{align}}
where {\small$\omega^{(t)}_i(k+1)$} and {\small$\omega^{(c)}_i(k+1)$} are drawn from {\small$\Omega_t \sim \mathcal{N}(0.00,0.005)$} and {\small$\Omega_c \sim \mathcal{N}(0.00,0.005)$} respectively.

\begin{figure}
  \centering
  \subfigure[]{\includegraphics[width=0.23\textwidth]{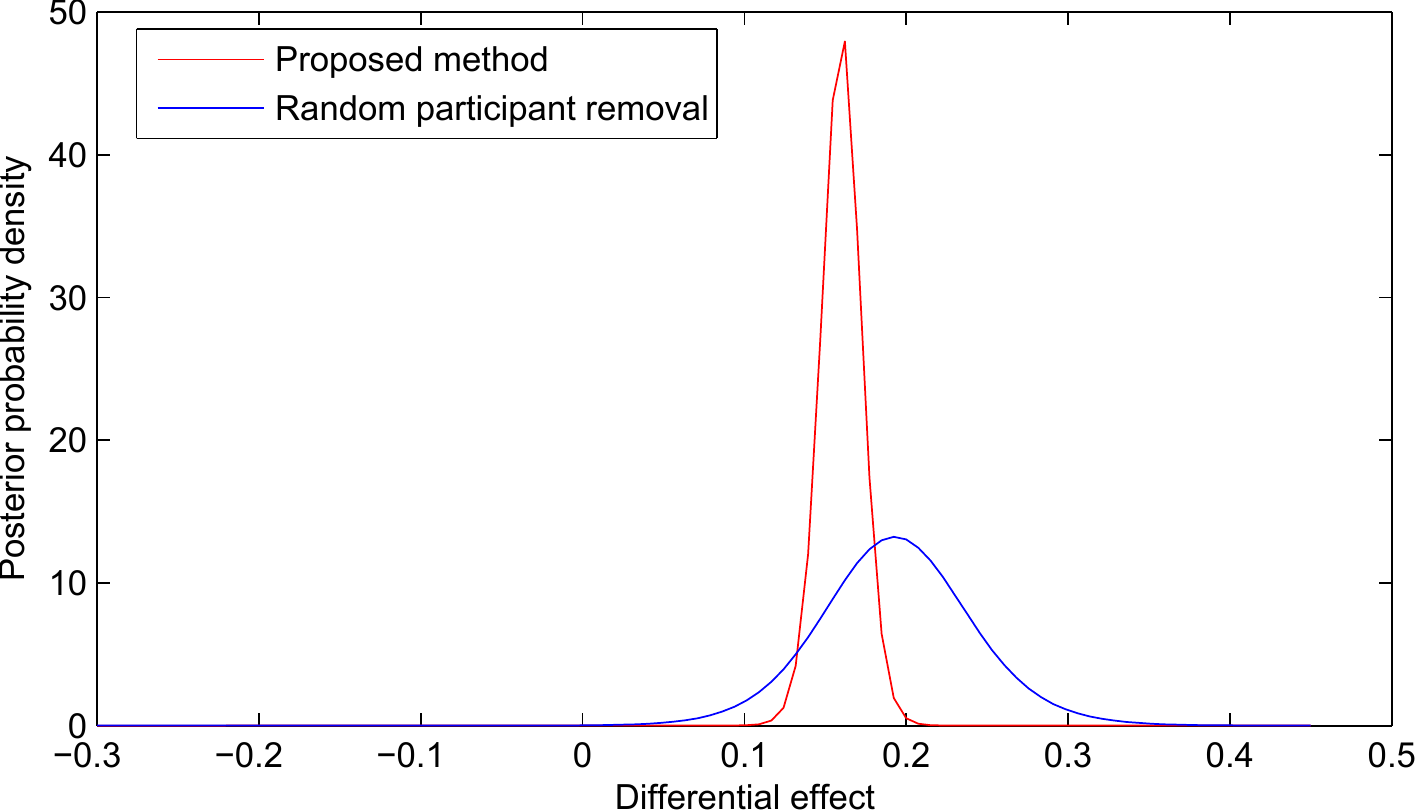}\label{f:posteriors}}~
  \subfigure[]{\includegraphics[width=0.23\textwidth]{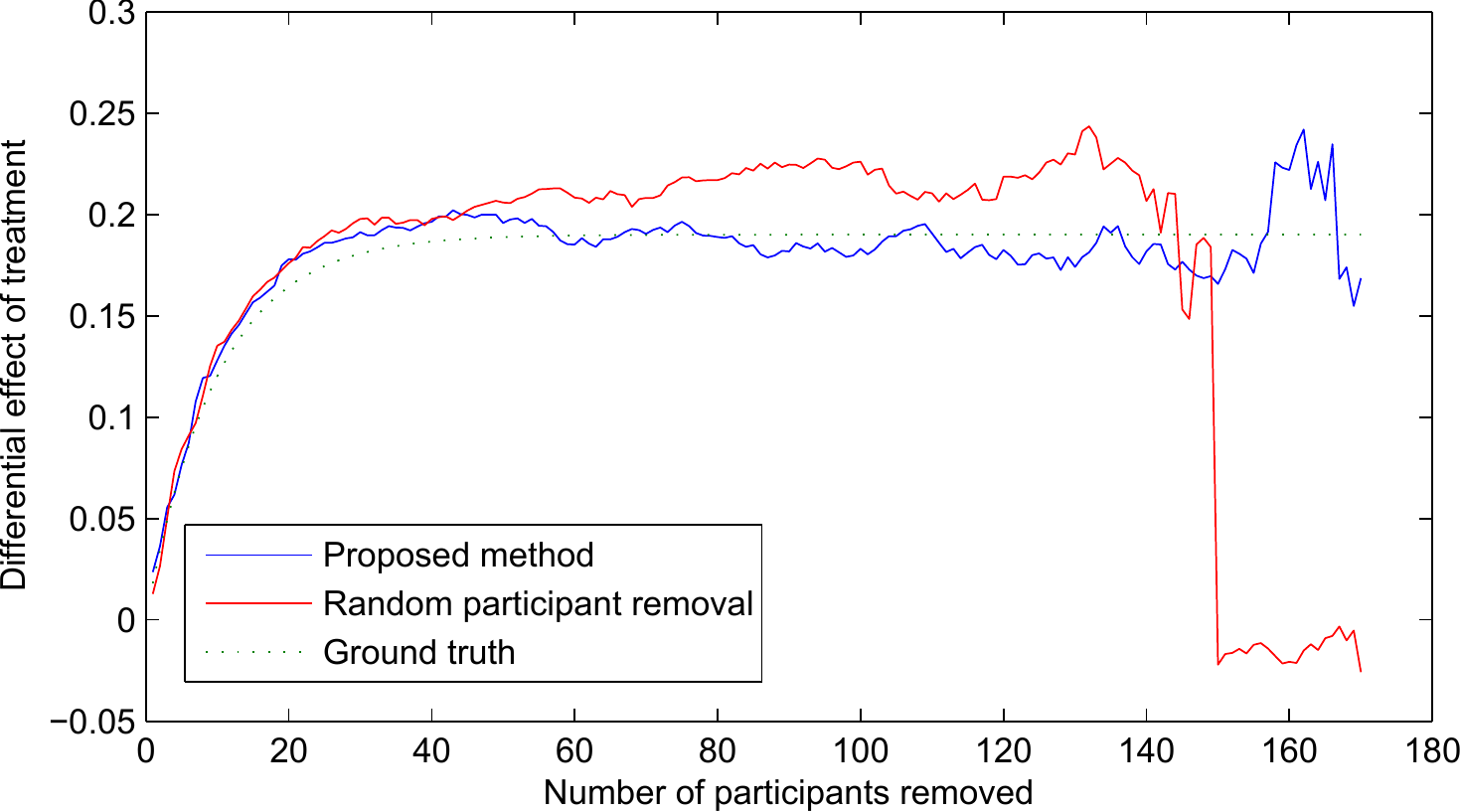}\label{f:mapPlot}}
  \subfigure[]{\includegraphics[width=0.23\textwidth]{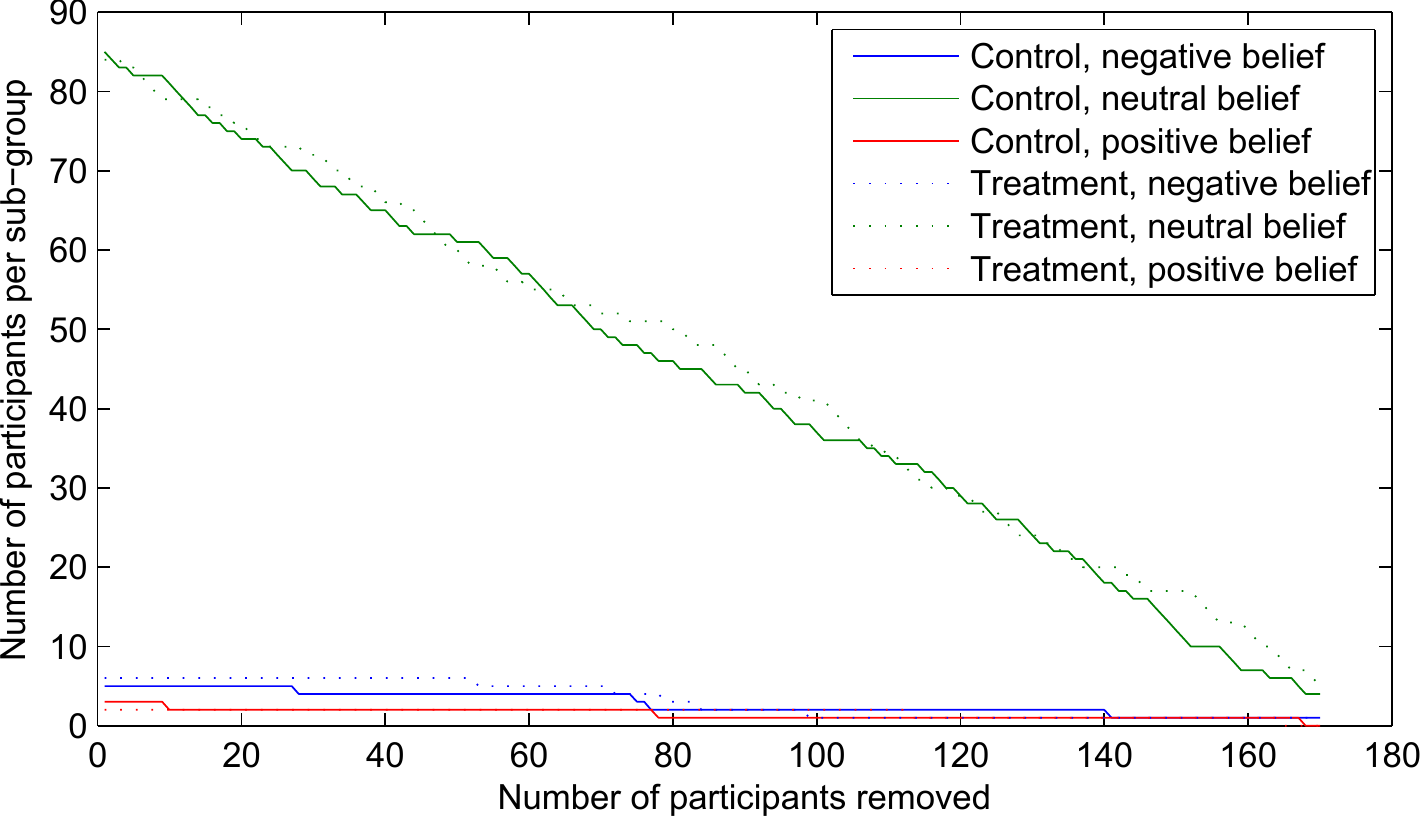}\label{f:groupSizesA}}~
  \subfigure[]{\includegraphics[width=0.23\textwidth]{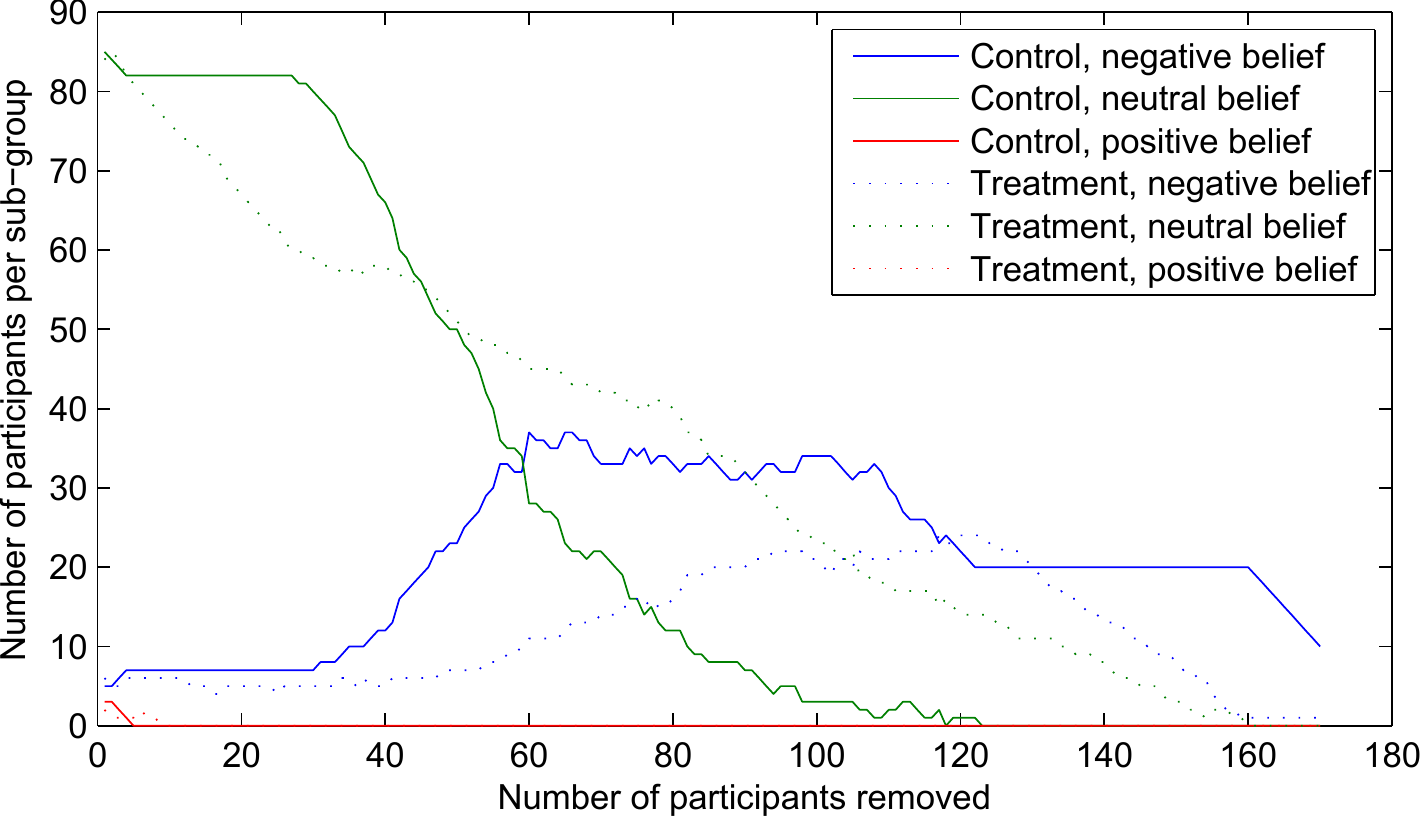}\label{f:groupSizesB}}
  \vspace{-10pt}
  \caption{\small (a) Posteriors of the differential effect of treatment after the removal of 120 participants; (b) the \emph{maximum a posteriori} estimates of the differential effect of treatment during the course of the trial;
  the changes in the sample sizes within each of the six participant sub-groups observed in our experiment using (c) random selection based participant removal, and (d) the proposed method.
  }
  \vspace{-10pt}
\end{figure}

\subsection{Results and discussion}
Using the same data obtained by simulating the experiment outlined in the previous section, we compared the proposed method with the current practice of randomly selecting participants which are to be removed from the trial. In both cases, data was analyzed using the Bayesian method proposed in~\citep{Aran2012g}. A typical result is illustrated in Fig.~\ref{f:posteriors}; the plot shows the posterior distributions of the differential effect of the treatment inferred after the removal of 120 individuals, obtained using the proposed method (red line) and random selection (blue line). The most notable difference between the two posteriors is in the associated uncertainties -- the proposed method results in a much more peaked posterior i.e.\ a much more definite estimate. In comparison, the posterior obtained using random selection is much broader, admitting a lower degree of certainty associated with the corresponding estimate.

The accuracy of two methods is better assessed by observing their behaviour over time. The plot in Fig.~\ref{f:mapPlot} shows the \emph{maximum a posteriori} estimates of the differential effect of treatment obtained using the two methods during the course of the trial. Also shown is the `ground truth', that is, the actual differential effect which we can compute exactly from the setup of the experiment. In the early stages of the trial, while the magnitude of the accumulated effect is small and the number of participants large, the two estimates are virtually indistinguishable, and they follow the ground truth plot closely. As expected, as the number of participants removed increases both estimates start to exhibit greater stochastic perturbations. However, both the accuracy (that is, the closeness to the ground truth) and the reliability (that is, the magnitude of stochastic variability) of the proposed method can be seen to show superior performance -- its \emph{maximum a posteriori} estimate follows the ground truth more closely and fluctuates less than the estimate obtained when random selection is employed instead. It is also important to observe the rapid degradation of performance of the random selection method as the number of remaining participants becomes small, which is not seen in the proposed method. This too can be expected from the theoretical argument put forward earlier -- the statistically optimal choice of the sub-group from which participants are removed ensures that the posterior is not highly dependent on a small number of samples which would make it highly sensitive to the change in sample size.

Lastly, it is interesting to observe the differences between the changes in the sample sizes within each sub-group using the two approaches. This is illustrated using the plots in Fig.~\ref{f:groupSizesA} and~\ref{f:groupSizesB}. As expected, when random participant removal is employed, the sizes of all sub-groups decrease roughly linearly (save for stochastic variability), as shown in Fig.~\ref{f:groupSizesA}. In contrast, the sub-group size changes effected by the proposed method show more complex structure, governed by the specific values of the belief and effect variables in our experiment. It is particularly interesting to note that the size changes are not only non-linear, but also non-monotonic. For example, the size of the control sub-group which includes individuals which correctly identified their group assignment (i.e.\ the sub-group {\small$G_{C-}$}) begins to increase notably after the removal of 30 participants and starts to decrease only after the removal of further 78 participants.

\section{Summary and conclusions}
We introduced a novel method for clinical trial adaptation by amending sample size. In contrast to all previous work in this area, the problem we considered was not \emph{when} sample size should be adjusted but rather \emph{which} particular samples should be removed. Our approach is based on the adopted stratification recently proposed for the analysis of trial outcomes in the presence of imperfect blinding. This stratification is based on the trial participants' responses to a generic auxiliary questionnaire that allows each participant to express belief concerning his/her intervention assignment (treatment or control). Experiments on a simulated trial were used to illustrate the effectiveness of our method and its superiority over the currently practiced random selection.

\begin{quote}
\begin{small}
\bibliographystyle{aaai}
\bibliography{../../../my_bibliography,../../../oa_physiology}
\end{small}
\end{quote}

\end{document}